\documentclass[cameraready]{Interspeech}
\title{Which Languages Transfer Best to Warlpiri? A Similarity-Based Study for Low-Resource ASR}

\author[affiliation={1}, orcid=0009-0005-3439-4970]{Pravina}{Mylvaganam}
\author[affiliation={1}]{Eliathamby}{Ambikairajah}
\author[affiliation={2}]{Ting}{Dang}
\author[affiliation={1}]{Vidhyasaharan}{Sethu}
\author[affiliation={3}]{Tuende}{Szalay}

\address{
    $^1$ University of New South Wales, Australia \\
    $^2$ University of Melbourne, Australia \\
    $^3$ University of Sydney, Australia 
}

\email{p.mylvaganam@unsw.edu.au , e.ambikairajah@unsw.edu.au, ting.dang@unimelb.edu.au, v.sethu@unsw.edu.au, tuende.szalay@sydney.edu.au}


\usepackage{comment}

\usepackage{caption}
\usepackage{subcaption}
\usepackage{graphicx}
\usepackage{multirow}
\usepackage{tablefootnote}
\usepackage{soul, xcolor}
\usepackage{amsmath, amsfonts}

\begin{document}

\maketitle

\begin{abstract}

This paper investigates how language similarity can improve cross-lingual transfer for automatic speech recognition (ASR) in extremely low-resource settings. Warlpiri, an Australian Aboriginal language, has very limited transcribed speech data, making transfer learning essential. We propose a framework combining acoustic similarity from pre-trained speech models with linguistic similarity based on typology, phoneme inventories, grammatical, and syntactic features to rank high-resource source languages and evaluate their effectiveness for ASR transfer to Warlpiri. Experiments with Whisper show that acoustically and typologically similar languages outperform monolingual and multilingual baselines. Assamese and Hindi achieve substantial reductions in word and character error rates. Correlation analysis further indicates that acoustic similarity is the strongest predictor of fine-tuning performance, while phoneme inventory and typological similarity better explain zero-shot transfer.
\end{abstract}

\section{Introduction}
\vspace{-5pt}

Automatic Speech Recognition (ASR) has become a transformative technology that enables natural human–computer interaction in applications such as virtual assistants, language learning, and accessibility tools. In recent years, ASR has advanced markedly due to large-scale annotated datasets and powerful deep learning models \cite{prabhavalkar2023end}. However, many languages, especially Indigenous and low-resource languages, remain underrepresented and perform worse than high-resource languages due to the limited transcribed corpora \cite{besacier2014automatic}. Warlpiri, an Australian Aboriginal language spoken by a small community in the Northern Territory \cite{bavin1993language}, exemplifies this challenge, as its extremely limited data hinder robust ASR development.


Cross-lingual transfer has emerged as a promising solution to data scarcity, leveraging acoustic and linguistic knowledge from high-resource languages to support low-resource ones \cite{das2015cross, khurana2024cross}. However, its effectiveness depends on the selection of suitable source languages for knowledge transfer \cite{malkin2022balanced}. Existing studies often choose source languages based on data availability, language families, or heuristics \cite{khare2021low, pandey2024towards, liu2024exploration, hou2021exploiting}, which may not reflect true acoustic or linguistic similarity \cite{gooskens2018mutual, heeringa2023comparing}. Recent work instead highlights acoustic and phonetic similarity as more reliable indicators \cite{umar2022investigating}, often yielding better transfer than genealogical or geographical closeness alone \cite{wu2021cross, kim2025improving}, given that geographically or genealogically related languages can still differ substantially in their acoustic properties \cite{themistocleous2022sonorant, harnud2021relation, lin2019choosing}. This highlights the need for systematic, similarity-based source language selection to maximize transfer in low-resource settings.

Warlpiri is typologically distant (\textit{i.e.,} differing substantially in grammar, morphology, and syntax) from most high-resource languages \cite{bavin1987warlpiri, shopen2001explaining, bryant2021contrasting} and has received limited attention, making it unclear which high-resource languages are the most suitable transfer sources. Its segmental and suprasegmental properties further underscore the need for a systematic similarity analysis: the vowel system of Warlpiri is relatively small, while its consonant inventory is extensive, including dental, alveolar, retroflex, and palatal sounds that are uncommon in many high-resource languages \cite{bavin1993language, bavin1987warlpiri, hale1983warlpiri}. Acoustically, it exhibits characteristic stress patterns, vowel harmony, and prosodic structures that differentiate it from languages such as English \cite{harvey2005vowel, pentland2005distinguishing, butcher2003acoustic}. These distinctive phonological and prosodic patterns suggest that transfer effectiveness is unlikely to be captured by broad criteria such as data availability alone, and instead depends on more fine-grained forms of similarity. This motivates a multi-level similarity analysis that examines syntactic, phoneme-inventory, grammatical, and typological similarity, with detailed acoustic measures, to systematically identify high-resource source languages that align with Warlpiri. 


A recent study \cite{ambikairajah2025study} examined similarity between Warlpiri and several high-resource languages using a language identification framework. While providing preliminary insights, it is limited to sentence-level acoustic similarity and does not consider fine-grained acoustic similarity across low- to high-level speech representations. More importantly, it excludes linguistic similarity measures, leaving the analysis largely descriptive. In addition, the proposed similarity metrics have not been evaluated on downstream tasks, leaving no empirical evidence of their utility for applications such as ASR improvements. 


To address these challenges, this study investigates two central questions: (1) which high-resource languages are most similar to Warlpiri across acoustic and linguistic dimensions, and (2) whether similarity-based source language selection improves the effectiveness of cross-lingual transfer for ASR tasks. Specifically, we consider two categories of similarity: (1) embedding-based acoustic similarity derived from pre-trained speech models, and (2) linguistic-feature-based similarity computed from syntactic, phoneme inventory, grammatical, and typological characteristics. These measures are used to rank the high-resource languages and identify promising source languages for transfer through cross-lingual ASR experiments.


To our knowledge, this is the first systematic study analysing similarity between Warlpiri and high-resource languages across acoustic and linguistic features, and linking it directly to ASR performance. It provides a principled approach for developing ASR systems for low-resource languages.



\section{Similarity Analysis}
\vspace{-5pt}
The detailed similarity analysis between Warlpiri and the selected high-resource languages is presented in this section. Candidate languages are first identified using the LID-based similarity approach \cite{ambikairajah2025study}, which selects languages proximate to Warlpiri, providing a principled basis for in-depth similarity analysis while ensuring computational feasibility by limiting the study to the most acoustically proximate languages.


\subsection{LID-based similarity}
\label{LID}
\vspace{-5pt}
This approach identifies acoustically similar high-resource languages to Warlpiri as suitable candidates for further similarity analysis using our proposed acoustic and linguistic feature methods (sections~\ref{AS} and \ref{LS}). We employ a widely used multilingual pre-trained ECAPA-TDNN model \cite{desplanques2020ecapa} trained on the diverse VoxLingua107 \cite{valk2021voxlingua107}, providing broad coverage across language families for reliable cross-language comparison. Warlpiri utterances are fed to the model to generate predictions over 107 languages, with higher predicted probabilities indicating greater acoustic similarity to Warlpiri \cite{ambikairajah2025study}.


\subsection{Acoustic Similarity}
\label{AS}
\vspace{-5pt}

We propose an acoustic approach to quantify language similarity using only speech signals. Speech embeddings are first obtained using the pre-trained speech model $f(\cdot)$. The model maps each speech utterance $x$ to a fixed-dimensional embedding space $\mathbf{e} = f(x) \in \mathbb{R}^d$ . The language similarity, $s$, is then calculated using the cosine similarity between the embedding vectors $\mathbf{e}_{w}$ and $\mathbf{e}_{\ell}$ of Warlpiri and a high-resource language $\ell$:

\vspace{-7pt}
\begin{equation}
\mathrm{s}(x^{w}, x^{\ell}) = (\mathbf{e}_{w} \cdot \mathbf{e}_{\ell}) / (\lVert \mathbf{e}_{w} \rVert \, \lVert \mathbf{e}_{\ell} \rVert)
\end{equation}

For each Warlpiri utterance, similarity is computed against a representative subset of utterances (described in Section~\ref{sec3.1}) from language $\ell$. The resulting pairwise cosine similarities are averaged to obtain a similarity score for language $\ell$ per Warlpiri utterance, and the overall acoustic similarity $S(w,\ell)$ is computed by averaging these scores across all Warlpiri utterances.

\vspace{-10pt}
\begin{equation}
S(w,\ell)
=
\frac{1}{N_w}
\sum_{i=1}^{N_w}
\left(
\frac{1}{N_\ell}
\sum_{j=1}^{N_\ell}
\mathrm{s}\big(x_i^{w}, x_j^{\ell}\big)
\right)
\end{equation}

\noindent where $N_w$ and $N_\ell$ are the number of utterances in Warlpiri and language $\ell$, respectively. Retaining utterance-level embeddings rather than collapsing them into a single global representation ensures a robust and fine-grained estimate of cross-language acoustic proximity.

To further analyse how acoustic similarity evolves across representation depth, we extract embeddings from intermediate layers of transformer-based speech models. Layer-wise similarity is then computed using the same cosine formulation, yielding $S_k(w,\ell)$ for each layer $k$. This enables us to examine how similarity patterns shift from lower-level acoustic representations to higher-level phonetic or linguistic abstractions.

In this study, we evaluate several pre-trained speech models, including ECAPA-TDNN \cite{desplanques2020ecapa}, wav2vec 2.0 \cite{baevski2020wav2vec}, and XLSR-53 \cite{conneau2020unsupervised}. ECAPA-TDNN captures speaker and language-discriminative acoustic characteristics, while wav2vec 2.0 is a self-supervised model trained on large-scale English speech that generates contextualised embeddings containing rich acoustic and phonetic information. Using both models, we extract the final-layer embeddings for each utterance. Since ECAPA-TDNN lacks a transformer hierarchy and monolingual wav2vec 2.0 is not well-suited for cross-lingual layer-wise analysis, we focus on XLSR-53 \cite{conneau2020unsupervised}, a multilingual extension of wav2vec 2.0 trained on 53 languages, for layer-wise evaluation. Unlike its monolingual counterpart, XLSR-53 is particularly well-suited for cross-lingual investigations. Its hierarchical transformer architecture produces representations that progress from low-level acoustic features to higher-level phonetic and linguistic abstractions \cite{pasad2021layer}, enabling a detailed examination of how Warlpiri’s similarity to other languages evolves across layers.


Finally, acoustic similarity analysis ranks high-resource languages by proximity to Warlpiri across models and layers. Although Warlpiri is unseen during training, embedding similarity serves as a proxy for acoustic and phonetic affinity.

\subsection{Linguistic-feature Similarity}
\label{LS}
\vspace{-5pt}

We further examined non-acoustic features, i.e., typological properties documented in linguistic studies \cite{ye2023study}. We analyze language similarity across four complementary dimensions: syntactic structure, phoneme inventory, grammatical features, and overall typology, providing a multi-level characterization grounded in established linguistic resources. Languages are represented as feature vectors extracted from publicly available databases, and similarity between Warlpiri and each high-resource language is computed using cosine and Hamming distances. The four dimensions are defined as follows:

\begin{enumerate}

    \item \textbf{Syntactic Distance: }Measures similarity between syntax feature vectors using data from the World Atlas of Language Structures (WALS) \cite{wals}, Syntactic Structures of World Language Structures (SSWL) \cite{sswl}, and Ethnologue \cite{Ethnologue2025}, encoding sentence structure and word order.

    \item \textbf{Inventory Distance: }Measures similarity in phoneme inventories, reflecting the sound systems of languages. The binary phoneme vectors are derived from the PHOIBLE database \cite{PHOIBLE2.0}. 


    \item \textbf{Grammatical Distance: }Captures grammatical structure, such as case marking, agreement, tense–aspect–mood, using Grambank feature vectors \cite{skirgaard2023grambank}.

    \item \textbf{Typology Distance: }Overall structural similarity from concatenated vectors of the above three dimensions, reflecting global typological similarity.
\end{enumerate}

To ensure reliability, features with missing values were ignored rather than predicted \cite{kim2025improving, skirgaard2023grambank}. Predicting missing feature values can introduce artificial patterns that distort similarity measurements, so only attested feature values were used to compute similarity scores.


\section{Experiments}
\label{sec3}

\subsection{Datasets}
\label{sec3.1}
\vspace{-5pt}
Warlpiri speech recordings and their corresponding transcriptions were obtained from the DoReCo dataset \cite{doreco-warl1254}. The recordings were first pre-processed by removing low-quality segments and utterances containing excessive noise or irrelevant speech. The remaining audio signals were then downsampled to 16 kHz to match the input requirements of the speech models used in this study. The final dataset consists of approximately 1.5 hours of speech from 18 speakers. This reflects realistic documentation conditions for endangered languages, where only limited amounts of transcribed speech are typically available. The data were split into training, validation, and test sets with durations of 1 hour, 15 minutes, and 15 minutes, respectively. 

For acoustic analysis (Section~\ref{AS}), a representative subset of each high-resource language is randomly sampled from the VoxLingua107 training split \cite{valk2021voxlingua107}, on which the pre-trained models were optimized, ensuring reliable acoustic and phonetic representations. For computational efficiency, each subset is limited to 2,500 utterances (10 hours) per language while preserving speaker diversity and broad phonetic coverage.

\subsection{ASR model and Implementation Details}
\vspace{-5pt}
For ASR on Warlpiri, we use the Whisper model \cite{radford2023robust}, a large-scale multilingual encoder–decoder trained on diverse speech data, known for strong cross-lingual transfer in low-resource ASR \cite{gete2025whispering, liu2024exploration}. Given the limited availability of Warlpiri speech data, we use the Whisper small variant for all experiments. It comprises a 12-layer encoder and decoder, a model dimension of 768, 12 attention heads, and 244 million trainable parameters. This configuration balances recognition performance and computational efficiency, ensuring consistent and controlled cross-lingual transfer experiments.


We start with the multilingual Whisper small model and fine-tune it individually on each selected high-resource source language dataset. The resulting source-adapted models are then fine-tuned on Warlpiri data. All encoder and decoder layers are updated, as full-model fine-tuning enables better adaptation of both acoustic and linguistic representations to the low-resource language. The procedure follows Hugging Face guidelines \cite{HF}. Specifically, each model is trained for 10 epochs with a batch size of 2, using a 10\% warm-up of total steps, after which the learning rate reaches $10^{-5}$ and decays linearly. This setup ensures stable and sufficient training on the low-resource dataset.

During training, checkpoints were saved every 200 steps, and each was evaluated on the validation set. The checkpoint with the lowest word error rate (WER) was selected as the final model. This configuration was kept consistent across all experiments to ensure fair comparison between source languages. The final model was evaluated on a held-out test set that was not used during training using WER and character error rate (CER). Performance was compared against three baselines: (1) monolingual training, where the same model architecture was trained using only Warlpiri data without adaptation or transfer from any other language, (2) multilingual transfer using the original Whisper and XLSR-53 models, and (3) models pre-trained on languages least similar to Warlpiri. This setup assesses whether similarity-based source language selection improves ASR performance.

\section{Results and Discussion}

\subsection{Candidate Language Selection}
\vspace{-5pt}
Based on the LID-based pre-selection in Section~\ref{LID}, the top nine of 107 languages were selected as acoustically similar candidates: Assamese, Hindi, Tamil, Telugu, Malayalam, Finnish, Māori, Javanese, and Swahili, as they show higher similarity than the others. Notably, these languages are neither part of the Pama–Nyungan language family to which Warlpiri belongs, nor geographically close, indicating that selecting source languages solely based on genealogical or geographical similarity may not be effective, which is commonly used in the literature for cross-lingual transfer. This further underscores the need for a more principled strategy for selecting source languages.


To ensure a comprehensive evaluation, the acoustically least similar languages, English and Japanese, were also included. This resulted in a total of 11 high-resource languages, selected from an initial pool of 107, which were used for both the embedding-based acoustic similarity and linguistic similarity analyses described in Sections~\ref{AS} and~\ref{LS}. These languages were selected to span a broad range of similarity levels and language families for controlled comparison. However, due to missing linguistic feature data for Assamese and Japanese, only syntactic and phoneme inventory distances could be analyzed for these two languages.

\subsection{Language Similarity}
\vspace{-5pt}

\begin{figure}[!t]
    \centering
    \begin{subfigure}[b]{0.485\linewidth}
        \centering
        \includegraphics[width=\linewidth]{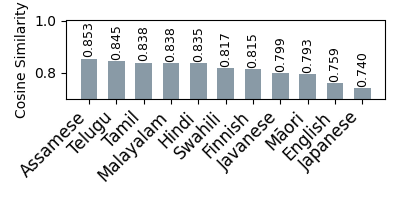}
        \caption{ECAPA-TDNN}
        \label{fig1_1}
    \end{subfigure}
    \hfill 
    \begin{subfigure}[b]{0.485\linewidth}
        \centering
        \includegraphics[width=\linewidth]{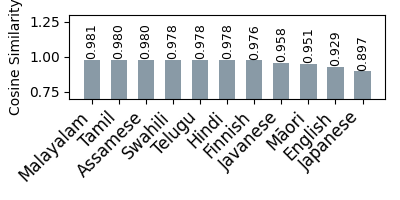}
        \caption{Wav2vec 2.0}
        \label{fig1_2}
    \end{subfigure}
    \vspace{-5pt}
    \caption{Embedding-based cosine similarity between Warlpiri and selected high-resource languages using ECAPA-TDNN and Wave2vec 2.0.}
    \label{fig1}
    \vspace{-20pt}
\end{figure}

\noindent\textbf{Embedding-based acoustic similarity: }Figure~\ref{fig1} shows cosine similarity between Warlpiri and 11 selected high-resource languages using final-layer embeddings from ECAPA-TDNN and wav2vec 2.0, as they are the standard representation for downstream fine-tuning. Although rankings vary slightly, a consistent pattern appears: Assamese, Tamil, Telugu, Malayalam, and Hindi rank highest, indicating stronger acoustic similarity to Warlpiri, while English and Japanese rank lower. This consistency across models suggests that the observed similarities reflect genuine acoustic relationships rather than model-specific effects.

\begin{figure}[h!]
\vspace{-10pt}
    \centering
    \includegraphics[width=0.8\linewidth]{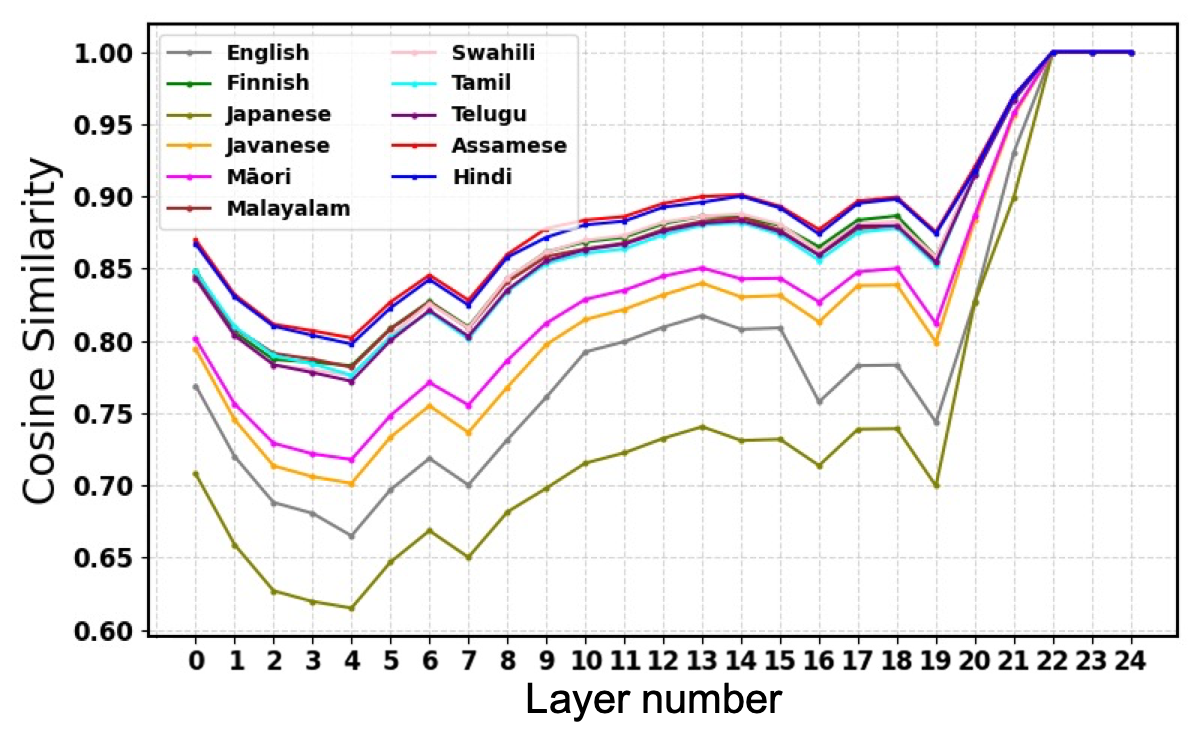}
    \vspace{-12pt}
    \caption{Cosine similarity between Warlpiri and selected high-resource languages across layers of XLSR-53.}
    \label{fig2}
    \vspace{-10pt}
\end{figure}

Figure~\ref{fig2} reinforces these findings with a layer-wise analysis using the XLSR-53 model, showing cosine similarity between Warlpiri and the 11 languages across the final convolutional layer (Layer 0) and transformer layers (Layer 1-24). Prior work \cite{pasad2021layer} indicates that lower layers capture acoustic features, middle layers encode phonetic information, and higher layers represent task-specific characteristics. This enables examination of how Warlpiri’s similarity to high-resource languages evolves across model layers. 
Among the 11 languages, Assamese and Hindi show consistently high similarity with Warlpiri across lower and middle transformer layers, 
indicating strong acoustic and phonetic proximity, whereas English and Japanese remain relatively dissimilar. Together, Figures~\ref{fig1} and \ref{fig2} show that Assamese and Hindi are acoustically closer to Warlpiri, supporting a similarity-based source language selection strategy.

\begin{figure}[t!]
    \centering
    \begin{subfigure}[b]{0.485\linewidth}
        \centering
        \includegraphics[width=\linewidth]{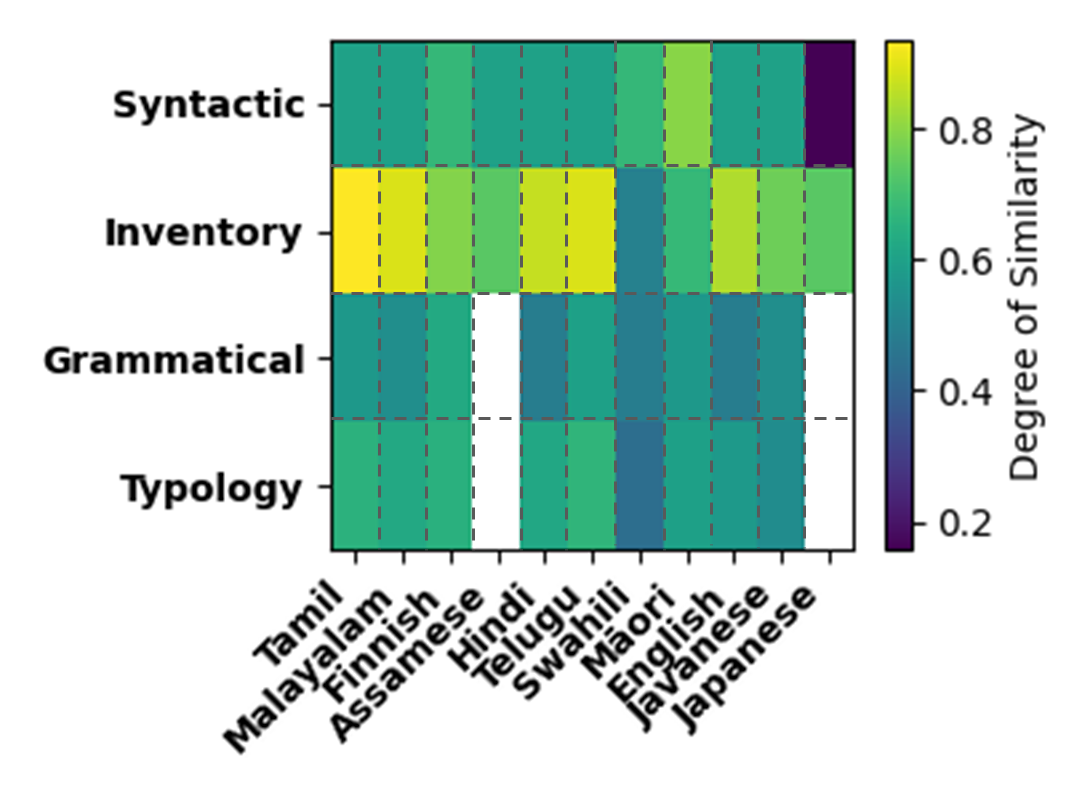}
        \label{fig3_1}
    \end{subfigure}
    \begin{subfigure}[b]{0.485\linewidth}
        \centering
        \includegraphics[width=\linewidth]{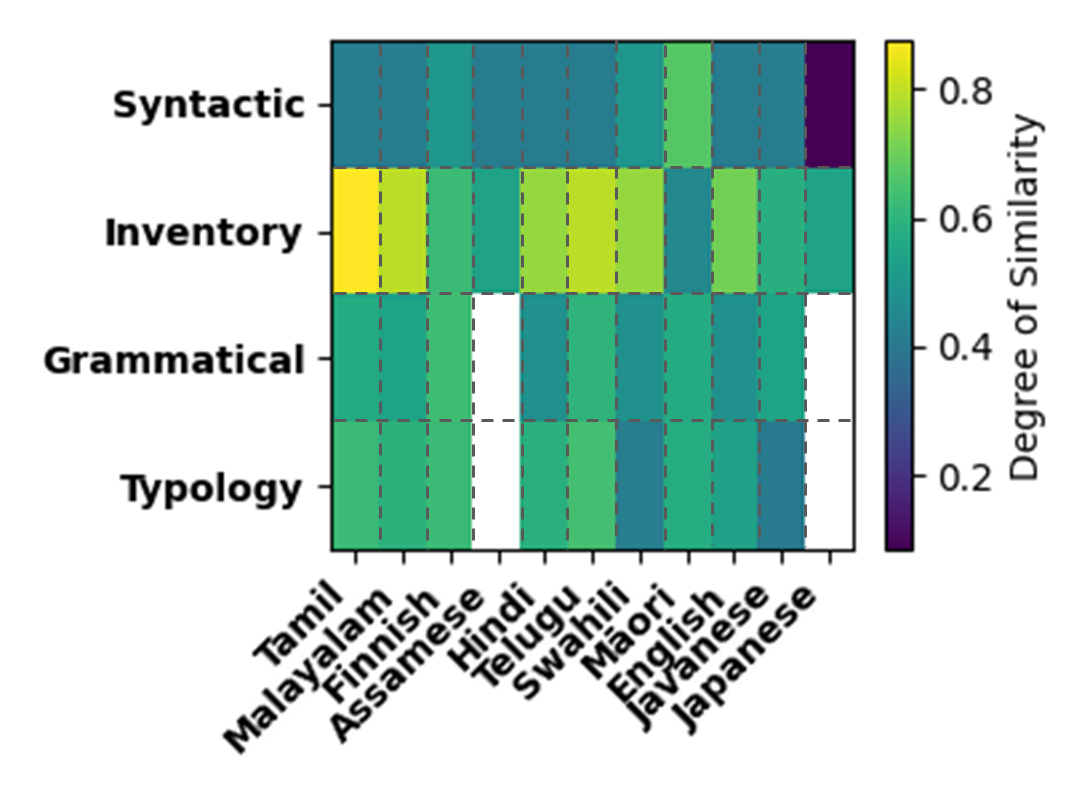}
        \label{fig3_2}
    \end{subfigure}
    \vspace{-20pt}
    \caption{The language similarity matrix measures cosine (left) and hamming (right) similarity between Warlpiri and selected high-resource languages.}
    \vspace{-18pt}
    \label{fig3}
\end{figure}

\noindent\textbf{Linguistic feature similarity: }Figure~\ref{fig3} shows the language similarity matrix based on linguistic features. Tamil, Telugu, Finnish, and Hindi exhibit greater overall linguistic similarity to Warlpiri. Assamese also ranks relatively high based on available features, while Japanese exhibits lower similarity.

The linguistic-feature results largely align with the embedding-based acoustic findings, suggesting that acoustically similar languages often share linguistic traits as well. However, some discrepancies exist; for instance, Māori shows higher syntactic similarity despite lower acoustic proximity, highlighting the importance of considering multiple similarity dimensions. By integrating acoustic and linguistic-feature analyses, we gain a more comprehensive view of language proximity between Warlpiri and the 11 high-resource languages, enabling the identification of robust candidate source languages for downstream speech tasks.


\subsection{ASR Performance}
\vspace{-5pt}
Table~\ref{tab:asr_backbone_results} presents the ASR performance of the five most similar and three least similar source languages selected from the 11 candidate high-resource languages based on different similarity measures. Among the baseline, the monolingual model performs poorly, with a WER of 86.9\% and a CER of 41.3\%, indicating that training solely on the limited Warlpiri data is insufficient. The multilingual Whisper model improves performance substantially, reducing WER and CER by 52.8\% and 63.4\% relative to the monolingual baseline, demonstrating the benefits of cross-lingual transfer. Fine-tuning the self-supervised XLSR-53 model, previously shown effective for low-resource languages \cite{arisaputra2024xls}, also improves results but struggles to predict Warlpiri accurately given the extremely limited data.

When Whisper models are fine-tuned using our similarity-based source selection, a clear trend emerges. Assamese, the most acoustically similar language to Warlpiri, achieves the best performance with a WER of 32.6\% and CER of 12.3\%, surpassing all baselines. Hindi, Telugu, and Tamil also perform competitively, with WERs between 37.6\% and 40.7\%, showing that acoustically and linguistically similar languages enable better transfer, due to shared vowel space and overlapping consonant inventories. In particular, Assamese shows comparable vowel distributions and bilabial/alveolar consonants relative to Warlpiri, while Tamil shares retroflex consonants, supporting their higher similarity scores. In contrast, less similar languages like Japanese and Javanese perform worse, with WERs of 49.7\% and 48.0\%, respectively. These results demonstrate that similarity-based source selection consistently improves cross-lingual ASR, with acoustically close languages like Assamese and Hindi providing the most effective transfer.


\begin{table}[t]
\footnotesize
\centering
\caption{ASR performance (WER and CER) using different backbones and source languages. Best results are highlighted in \textbf{Bold}.}
\label{tab:asr_backbone_results}
\vspace{-10pt}
\begin{tabular}{lccc}
\toprule
Backbone & Source languages & WER (\%) & CER (\%) \\
\midrule
\multirow{3}{*}{Whisper} & \textit{Baselines} &  &  \\
 & Monolingual   & 86.9 & 41.3 \\
 & Multilingual  & 41.0 & 15.1 \\
XLSR-53 & -- & 72.7 & 26.5 \\
\midrule
\multirow{8}{*}{Whisper} & Assamese & \textbf{32.6} & \textbf{12.3} \\
 & Hindi  & {37.6} & 14.3 \\
 & Telugu & 39.9 & 15.6 \\
 & Tamil & 40.7 & 15.2 \\
 & English   & 41.1 & 15.7 \\
 & Finnish   & 41.5 & 15.8 \\
 & Javanese & 48.0 & 24.1 \\
 & Japanese  & 49.7 & 24.5 \\
\bottomrule
\end{tabular}
\vspace{-10pt}
\end{table}

\subsection{Correlation between Similarity and ASR Performance}
\vspace{-5pt}
To evaluate the effectiveness of each language similarity measure for cross-lingual ASR, we computed the Spearman rank correlation between the similarity scores and the resulting WER and CER values under two settings: (1) zero-shot, where source-adapted models are evaluated directly on Warlpiri speech without any further training, and (2) fine-tuning, where models are adapted using our similarity-based approach. Table~\ref{tab1} summarizes the strength of the relationship between each similarity measure and ASR performance in these scenarios. In this analysis, a strong negative correlation indicates that higher similarity is associated with lower error rates, making it a reliable predictor of cross-lingual transfer.

In the zero-shot setting, inventory similarity shows the strongest correlation overall, followed by typology and acoustic measures, indicating their importance for direct cross-lingual transfer without adaptation. In the fine-tuning setting, acoustic similarity demonstrates the highest correlation, followed by typology, suggesting that once adaptation is allowed, acoustic proximity between the source and target languages becomes the dominant factor for improving ASR performance. By contrast, syntactic and grammatical similarity measures show weaker or inconsistent correlations with ASR metrics in both settings. Together, these findings imply that structural compatibility, particularly phoneme inventory overlap, is more important when no adaptation data is available (zero-shot), whereas acoustic proximity dominates once fine-tuning is introduced in cross-lingual transfer for Warlpiri in this study.


\begin{table}[]
\footnotesize
\centering
\caption{Spearman correlation between language similarity and ASR performance (WER and CER) for zero-shot and fine-tuning settings. Best results are highlighted in \textbf{Bold}.}
\label{tab1}
\vspace{-10pt}
\begin{tabular}{lcc|cc}
\toprule
\multirow{2}{*}{Similarity} & \multicolumn{2}{c|}{Zero-shot} & \multicolumn{2}{c}{Fine-tuning} \\
\cline{2-5}
 & $\rho_\text{WER}$ & $\rho_\text{CER}$ & $\rho_\text{WER}$ & $\rho_\text{CER}$ \\
\hline
Acoustic     & \textbf{-0.45} & -0.12 & \textbf{-0.67} & \textbf{-0.62} \\
Syntactic    &  0.11 &  0.33 & -0.11 &  0.11 \\
Inventory    & \textbf{-0.54} & -0.35 & -0.31 &  0.24 \\
Grammatical  & -0.03 & -0.32 &  0.17 & -0.03 \\
Typology     & \textbf{-0.49} & \textbf{-0.54} & \textbf{-0.54} & \textbf{-0.49} \\
\bottomrule
\end{tabular}
\vspace{-10pt}
\end{table}

\vspace{-5pt}
\section{Conclusion}
\vspace{-5pt}
This study examined the role of language similarity in cross-lingual ASR for Warlpiri, an extremely low-resource Australian Aboriginal language. We presented a systematic framework that combines embedding-based acoustic similarity with linguistic-feature-based similarity to guide source language selection. Experimental results show that source languages selected based on acoustic and typological similarity consistently outperform monolingual and multilingual baselines. In particular, acoustically similar languages such as Assamese and Hindi provide the strongest transfer performance, while less similar languages perform substantially worse.

Correlation analysis shows that acoustic similarity is the strongest predictor of ASR performance in fine-tuning scenarios, while phoneme inventory and typological similarity are more informative for zero-shot transfer. This framework offers a practical and general strategy for improving ASR in extremely low-resource and endangered language settings.

\section{Acknowledgment}
The authors would like to thank the School of Electrical Engineering and Telecommunications at UNSW Sydney, Australia, for providing funding for this research initiative. 
Ethics approval was granted by the Human Research Ethics Advisory Panel Executive (HREAP) at UNSW Sydney, Australia, under Reference Number iRECS6867.

\section{Generative AI use disclosure}
During the preparation of this work, the authors used an AI tool in order to refine the academic language, improve the structural flow of the manuscript, and optimise technical terminology. After using this tool, the authors reviewed and edited the content as needed and take full responsibility for the content of the publication.

\bibliographystyle{IEEEtran}
\bibliography{ref}

\end{document}